\documentclass[11pt,a4paper]{article}
\usepackage{multirow}
\usepackage{rotating}
\usepackage{acl2016}
\usepackage{times}
\usepackage{latexsym}
\usepackage{graphicx}
\usepackage{amsmath}
\usepackage{amssymb}
\usepackage{url}
\usepackage[font={small}]{caption}

\def\newparagraph#1{\textbf{#1}}
\def\dnrm#1{\mbox{$_{\hbox{\scriptsize #1}}$}}

\aclfinalcopy 
\def\aclpaperid{72} 

\title{Single-Model Encoder-Decoder with Explicit
  Morphological Representation for Reinflection}

\author{Katharina Kann and Hinrich Sch\"utze\\
	    Center for Information \& Language Processing\\
	    LMU
	    Munich, Germany\\
	    {\tt kann@cis.lmu.de}}

\date{}

\newcounter{notecounter}

\newcommand{\enoteson}{\long\gdef\enote##1##2{{
\stepcounter{notecounter}
\large\bf
\hspace{1cm}\arabic{notecounter} $<<<$ ##1: ##2
$>>>$\hspace{1cm}}}}
\enoteson

\begin{document}

\maketitle

\begin{abstract}
Morphological reinflection is the task of generating a
target form given a source form, a source tag and a target
tag.  We propose a new way of modeling this task with neural
encoder-decoder models.  Our approach reduces
the amount of required training data for this architecture
and achieves state-of-the-art results, making
encoder-decoder models applicable to morphological
reinflection even for low-resource languages.  We further
present a new automatic correction method for the outputs
based on edit trees.
\end{abstract}

\section{Introduction}

Morphological analysis and generation of previously unseen
word forms is a fundamental problem in many areas of natural
language processing (NLP).  Its accuracy is crucial for the
success of downstream tasks like machine translation and
question answering.  Accordingly, learning morphological
inflection patterns from labeled data is an important
challenge.

The task of morphological reinflection (MRI) consists of
producing an inflected form for a given source form, source
tag and target tag. A special case is morphological
inflection (MI), the task of finding an inflected form for a
given lemma and target tag. An English example is
``tree''+PLURAL $\rightarrow$ ``trees''.
Prior work on
MI and MRI
includes machine learning models and models that exploit
the paradigm structure of the language \cite{ahlberg2015paradigm,dreyer2011non,nicolai2015inflection}.

In this work, we propose the neural encoder-decoder MED --
Morphological Encoder-Decoder -- a character-level
sequence-to-sequence attention model that is a
language-independent solution for MRI.  In contrast to prior
work, we train a \emph{single} model that is trained on  all
source to target mappings of the language that are attested
in the training set.  This radically reduces the
amount of training data needed for the encoder-decoder
because most MRI patterns occur in many source-target tag
pairs. In our model design, what is learned for one pair can be
transferred to others.

The key enabler for this single-model approach is a novel
representation we use for MRI. We encode the input as a
single sequence of (i) the morphological tags of the source
form, (ii) the morphological tags of the target form and (iii)
the sequence of letters of the source form. The output is
the sequence of letters of the target form. 
As the decoder produces each letter, the attention mechanism
can focus on \emph{the input letter sequence} for parts of the
output that simply copy the input. For other parts of the
output, e.g., an inflectional ending that is
predicted using the target tags, the attention mechanism
can focus on \emph{the target morphological tags}. In more
complex cases, \emph{simultaneous attention can be paid to subsequences of all three input types} --
source tags, target tags and input letter sequence.
We can train a
single generic encoder-decoder per language on this
representation that can handle all tag pairs, thus making it
possible to make efficient use of the available training
data.  MED outperformed other systems on the SIGMORPHON16
shared
task\footnote{\url{ryancotterell.github.io/sigmorphon2016/}}
for all ten languages that were covered \cite{kann16med,cotterell-sigmorphon2016}.

We also present POET -- Prefer Observed Edit Trees -- a new generic  method 
for 
correcting the output 
of an MRI system.
The
combination of MED and POET is state-of-the-art or close to
it on a CELEX-based evaluation of MRI even though this evaluation
makes it difficult to exploit generalizations across tag
pairs.

\section{Model Description}
\label{sec:model}

\newparagraph{Neural network model.} 
Our model is based on the
network architecture proposed by \newcite{bahdanau2014neural}
for machine translation.\footnote{Our implementation of MED
  is based on \url{github.com/mila-udem/blocks-examples/tree/master/machine_translation}.} They 
describe the model in detail; unless we explicitly say so in
the description of our model below, we use the same
network configuration as \newcite{bahdanau2014neural}.

\newcite{bahdanau2014neural}'s model
is an extension of the recurrent neural network (RNN) encoder-decoder developed by Cho et al. \shortcite{cho2014properties} and Sutskever et al. \shortcite{sutskever2014sequence}. 
The encoder of the latter consists of an RNN that reads an input sequence of 
vectors $x$ and encodes it into a fixed-length context vector $c$, computing hidden states $h_t$ and $c$ by
\begin{equation}
\label{eq:1}
  h_t = f(x_t, h_{t-1}), \quad  c = q({h_1, ..., h_{T_x}})
\end{equation}
with nonlinear functions $f$ and $q$. The decoder is trained to predict each output $y_{t}$ dependent
on $c$ and  previous predictions {$y_1$, ..., $y_{t-1}$}:
\begin{equation}
\label{eq:2}
  p(y) = \prod_{t=1}^{T_y} p(y_t | \{y_1, ..., y_{t-1}\}, c)
\end{equation}
with $y = (y_1, ..., y_{T_y})$ and each conditional probability being modeled with an RNN as
\begin{equation}
  p(y_t | \{y_1, ..., y_{t-1}\}, c) = g(y_{t-1}, s_t, c)
\end{equation}
where $g$ is a nonlinear function and $s_t$ is the hidden state of the RNN.

Bahdanau et al. \shortcite{bahdanau2014neural} proposed an
attention-based extension of this model that allows
different vectors $c_t$ for each step by automatic learning
of an alignment model. Additionally, they made the encoder
bidirectional: each hidden state $h_j$ at time
step $j$ does not only depend on the preceding, but also on
the following input:
\begin{equation}
  h_j = \left[\overrightarrow{h_j^T}; \overleftarrow{h_j^T}\right]^T
\end{equation}
The formula for $p(y)$ changes as follows:
\begin{align}
  p(y) &= \prod_{t=1}^{T_y} p(y_t | \{y_1, ..., y_{t-1}\}, x) \\
       &= g(y_{t-1}, s_t, c_t)
\end{align}
with $s_t$ being an RNN hidden state for time $t$ and $c_t$ being the weighted sum of the annotations $(h_1, ..., h_{T_x})$ produced by the encoder, using the attention weights.
Further descriptions can be found in \cite{bahdanau2014neural}.

The final model is a multilayer network with
a single maxout \cite{goodfellow2013maxout} hidden layer that
computes the conditional probability of each element in the
output sequence (a letter in our case,
\cite{pascanu2013construct}).  
As
MRI is less complex than  machine translation,
 we reduce the
number of hidden units and 
embedding size.  After initial experiments, we
fixed the hyperparameters of our system and did not further
adapt them to a specific task or language. Encoder and
decoder RNNs have 100 hidden units each. 
For training,
we use  stochastic gradient descent,
Adadelta \cite{zeiler2012adadelta} and
a minibatch size of 20. We initialize all weights in the encoder, decoder and the embeddings
except for the GRU weights in the decoder with the identity matrix as well as all biases with zero
\cite{le2015simple}. 
We train all models for 20,000 iterations. We
settled on this number in early experimentation because
training usually converged before that limit.

MED is an ensemble of five RNN encoder-decoders.
The final decision is made by majority voting.
In case of a tie, the
answer is chosen randomly among the most frequent predictions.

\newparagraph{Input and output format.} We define the alphabet
$\Sigma\dnrm{lang}$ as the set of characters used in the
application language. 
As each morphological tag consists of one or more subtags, e.g. ``number`` or ``case``, we further define 
$\Sigma\dnrm{src}$ and $\Sigma\dnrm{trg}$ as the set of morphological subtags seen during training as part of the source tag and target tag, respectively. Let $S\dnrm{start}$ and $S\dnrm{end}$ be  predefined start and end symbols.
Then each input of our system is of the format $S\dnrm{start} \Sigma\dnrm{src}^+ \Sigma\dnrm{trg}^+ \Sigma\dnrm{lang}^+ S\dnrm{end}$. In the same way, we define the output
format as $S\dnrm{start} \Sigma\dnrm{lang}^+ S\dnrm{end}$.

\begin{figure}
  \centering
  \includegraphics[width=0.4\textwidth]{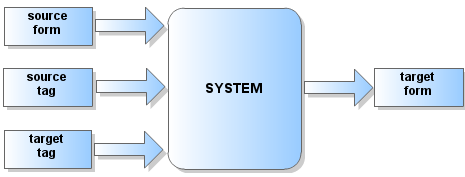}
  \caption{Overview of MED}
  \label{fig:systemoverview}
\end{figure}

A sample input for German is \textit{\textless w\textgreater~ IN=pos=ADJ IN=case=GEN IN=num=PL OUT=pos=ADJ OUT=case=ACC OUT=num=PL i s o l i e r t e r ~\textless/w\textgreater}. The system should produce the corresponding
output \textit{\textless w\textgreater~ i s o l i e r t e ~\textless/w\textgreater}.
The high-level structure of MED can be seen in Figure \ref{fig:systemoverview}.

\newparagraph{POET.}
We now describe 
POET (Prefer Observed Edit Trees), a new generic  
method for 
correcting the output 
of an MRI system. We use it in combination with MED in this
paper, but it can in principle be applied to any MRI system.

\begin{figure}
  \centering
  \includegraphics[width=0.3\textwidth]{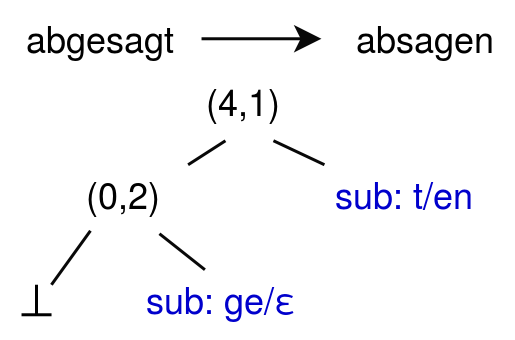}
  \caption{Edit tree for the inflected form
    \textit{abgesagt} ``canceled'' and its lemma
    \textit{absagen} ``to cancel''. The highest node
    contains the length of the parts before and after the
    LCS. The left node in the second row contains the length
    of the parts before and after the LCS of \textit{abge}
    and \textit{ab}. The prefix \textit{sub} indicates that
    the node is a substitution operation.}
  \label{fig:edittree}
\end{figure}

An edit tree $e(\sigma, \tau)$ specifies a transformation from a
source string $\sigma$ to a target string $\tau$
\cite{chrupala2008towards}. To compute $e(\sigma,\tau)$, we first
determine the longest common substring (LCS)
\cite{gusfield1997algorithms} between $\sigma$ and $\tau$ and then
recursively model the prefix and suffix pairs of the LCS.
If the length of LCS is zero for $(\sigma,\tau)$, 
then $e(\sigma,\tau)$ is simply the 
substitution operation that replaces $\sigma$ with $\tau$.  Figure
\ref{fig:edittree} shows an example.

Let $X$ be a training set for MRI. For each pair
$(s,t)$ of tags, we define:
\[
E_{s,t}  \!=\!  \{ e' | \exists x \!\in\! X: e'\!=\!e(x),s\!=\!S(x),t\!=\!T(x) \}
\]
where $S(x)$ and $T(x)$ are  source and target tags of $x$
and $e(x)$ is $e(\sigma(x),
\tau(x))$, the edit tree that transforms
the source form into the target form.

Let $\rho$ be a target form predicted by the MRI system for the source
form $\sigma$
and let $s$ and $t$ be source and target tags. POET 
does not change $\rho$ if $e(\sigma,\rho) \in E_{s,t}$.
Otherwise it 
replaces $\rho$ with $\tau$:
\begin{equation*}
\tau := 
\left\{ \begin{array}{ll}
    \tau' & \mbox{if} \ \ e(\sigma,\tau') \in E_{s,t},|\rho,\tau'|=1\\
    \rho & \mbox{else} \end{array} \right.
\end{equation*}
where $|\rho,\tau'|$ is
the Levenshtein distance.
If there are several forms $\tau'$ with edit distance 1, we
select the one with the most frequent edit tree. Ties are broken randomly.

We observed that MED sometimes makes errors that are close
to the target, but differ by one edit operation. Those
errors are often not covered by edit trees that are
observed in the training data whereas the correct form is.
Thus, substituting a form not supported by an observed edit
tree with a close one that is supported promises to reduce the error rate.

The effectiveness of POET depends on a training set that is
large enough to cover the possible edit trees that can occur
in reinflection in a language. Thus, if the training set is
not large enough in this respect, then POET will not be
beneficial.

\section{Experiments}\label{sec:experiments}
We compare MED with the three models of
\newcite{dreyer2008latent} as well as with two recently
proposed models: (i) discriminative string transduction
\cite{durrett2013supervised,nicolai2015inflection}, the
SIGMORPHON16 baseline,  and (ii)
\newcite{DBLP:journals/corr/FaruquiTND15}'s encoder-decoder
model.\footnote{For our experiments we ran the code
  available at \url{github.com/mfaruqui/morph-trans}. We
  used the \textit{enc-dec-attn} model as overall results
  for the CELEX task were better than with the
  \textit{sep-morph} model.}  We call the latter MODEL*TAG
as it requires training as many models as there are target
tags.

We evaluate MED on two MRI tasks: CELEX and SIGMORPHON16.

\newparagraph{CELEX.}
This task is based on complete inflection tables for German
extracted from CELEX.  For this experiment we follow
\newcite{dreyer2008latent}. We use four pairs of
morphological tags and corresponding word forms from the
German part of the CELEX morphological database.  The 4
different transduction tasks are: 13SIA $\rightarrow$ 13SKE,
2PIE $\rightarrow$ 13PKE, 2PKE $\rightarrow$ z and rP
$\rightarrow$ pA.\footnote{13SIA=1st/3rd sg.\ ind.\ past;
  13SKE=1st/3rd sg.\ subjunct.\ pres.; 2PIE=2nd
  pl.\ ind.\ pres.; 13PKE=1st/3rd pl.\ subjunct.\ pres.;
  2PKE=2nd.\ pl.\ subjunct.\ pres.; z=infinitive; rP=imperative
  pl.; pA=past part.}  An example for this task would be to
produce the output \textit{gesteuert} (target tag \textit{pA}) for the source
\textit{steuert} (source tag \textit{rP}).
 To do so, the system has to learn that the
prefix \textit{ge-}, which is used for many participles in
German, has to be added to the beginning of the original
word form.

We use the same data splits as \newcite{dreyer2008latent}, dividing the original $2500$ samples for each tag into five folds, each consisting 
of $500$ training and $1000$ development and $1000$ test
samples. We train a separate model for each fold and report
exact match accuracy, averaged over the five folds, as our final result.

\newparagraph{SIGMORPHON16.}
This task covers eight languages and does not provide
complete paradigms, but only a set of quadruples, each
consisting of word form, source tag, target tag and target
form.  The main difference to CELEX is that the number of
tag pairs is large, resulting in much less training
data per tag pair.  The number of tag pairs varies by
language with Georgian being an extreme case; it has 28 tag
pairs in dev that appear less than 10 times in train.  For
each language, we have around 12,800 training and 1600 development
samples. We report exact match accuracy on the development set, as the final test data of the shared task is not 
publically available yet.

\section{Results}
Table \ref{table:CELEXresults} gives CELEX results. MED+POET
is better than prior work on one task, close in
performance on two and worse by a small amount on the third.
Unlike \newcite{dreyer2008latent}'s models,
MED does not use any 
hand-crafted features.
MED's results are
weakest on 13SIA. Typical errors on this task include 
epenthesis (e.g., \textit{zirkle} vs.\  \textit{zirkele}) 
and irregular
verbs (e.g., \textit{abhing} vs.\ 
\textit{abh\"{a}ngte}).

\begin{table}
\centering
\begin{tabular}{ll|cccc} 
& {model} & \rotatebox{90}{13SIA} & \rotatebox{90}{2PIE} & \rotatebox{90}{2PKE} & \rotatebox{90}{rP} \\
 \hline
    \multirow{3}{*}{\footnotesize\rotatebox{90}{Dreyer}}& backoff & 82.8 & 88.7 & 74.7 & 69.9 \\ 
& lat-class & 84.8 & 93.6 & 75.7 & 81.8 \\ 
& lat-region & \textbf{87.5} & 93.4 & 87.4 & \textbf{84.9} \\\hline
& baseline & 77.6 & \textbf{95.1} & 82.5 & 69.6\\
& MODEL*TAG & 76.4 & 92.1 & 83.4 & 81.8 \\
& MED & 82.3 & 94.4 & 86.8 & 83.9 \\
& MED+POET & 83.9 & 95.0 &  \textbf{87.6} & 84.0 
 \end{tabular}
\caption{Exact match accuracy of MRI on CELEX. Results of
  \cite{dreyer2008latent}'s model are from their paper;
  backoff: \textit{ngrams+x} model; lat-class:
  \textit{ngrams+x+latent class} model; lat-region: 
\textit{ngrams+x+latent class+latent region} model;
baseline: SIGMORPHON16 baseline.}
\label{table:CELEXresults}
\end{table}

For SIGMORPHON16, Table \ref{table:SIGMORPHONresults} shows
that MED outperforms the baseline for all eight languages.
Absolute performance and variance is probably influenced by
type of morphology (e.g., templatic vs.\ agglutinative),
regularity of the language, number of different tag pairs
and other factors. 
MED performs well even for complex and diverse languages like
Arabic, Finnish, Navajo and Turkish, suggesting that 
the type of attention-based encoder-decoder we use -- single-model,
using an explicit morphological representation --
is a good choice for MRI.

We do not compare to MODEL*TAG here because it requires
training a large number of individual networks.
This is a
disadvantage compared to MED both in terms of the number of
models that need to be trained and in terms of the effective
use of the small number of training examples that are
available per tag pair.

\begin{table}
\centering
\begin{tabular}{l|cccc} 
&&\multicolumn{2}{c}{MED}\\
  & baseline &  average& ensemble\\\hline
 {Arabic} & 58.8 & 83.1 (0.4)& \textbf{88.8} \\
 {Finnish} & 64.6 & 92.5 (0.8)& \textbf{95.6} \\
 {Georgian} & 91.5 & 95.7 (0.3)& \textbf{97.3} \\
 {German} & 87.7 & 92.1 (0.5)& \textbf{95.1} \\
 {Navajo} & 60.9 & 85.0 (1.1)& \textbf{91.1} \\
 {Russian} & 85.6 & 84.2 (0.3)& \textbf{88.4} \\
 {Spanish} & 95.6 & 96.3 (0.3)& \textbf{97.5} \\
 {Turkish} & 54.9 & 94.7 (1.3)& \textbf{97.6} 
 \end{tabular}
\caption{Exact match accuracy of MRI on SIGMORPHON16; baseline:
SIGMORPHON16 baseline; MED/average: average of five MED
models (standard deviation in parentheses); MED/ensemble: majority voting of five MED models.}
\label{table:SIGMORPHONresults}
\end{table}

\textbf{POET} improves the results for all tag pairs for
CELEX. 
However, 
initial experiments indicated that it is not effective for
SIGMORPHON16 because its training sets are not large enough.

\section{Analysis}\label{sec:analysis}
The main innovation of our work is that MED learns a single
model of all MRI patterns of a language and thus can
transfer what it has learned from one tag pair to another
tag pair. Using CELEX, we now analyze how much our design contributes to
better performance by conducting 
two experiments in which we gradually decrease the training set
in two different ways.
(i)
Large general training set. We only reduce the number of
training examples available for a tag pair $(s,t)$ and retain all other
training examples. (ii) Small training set. 
We  reduce the number of
training examples available for all tag pairs, not just for one.

A typical example of the large general training set scenario
is that familiar second person
forms are rare in 
genres
like encyclopedia and news. So a training set derived from
these genres will be large, but it will have very few 
tag pairs whose target tag is  familiar second person.

A typical example of the small training set scenario is that
we are dealing with a low-resource language.

In the following two experiments,
we only reduce the training set and do not change
the test set.

\begin{figure}
  \centering
  \includegraphics[width=0.35\textwidth]{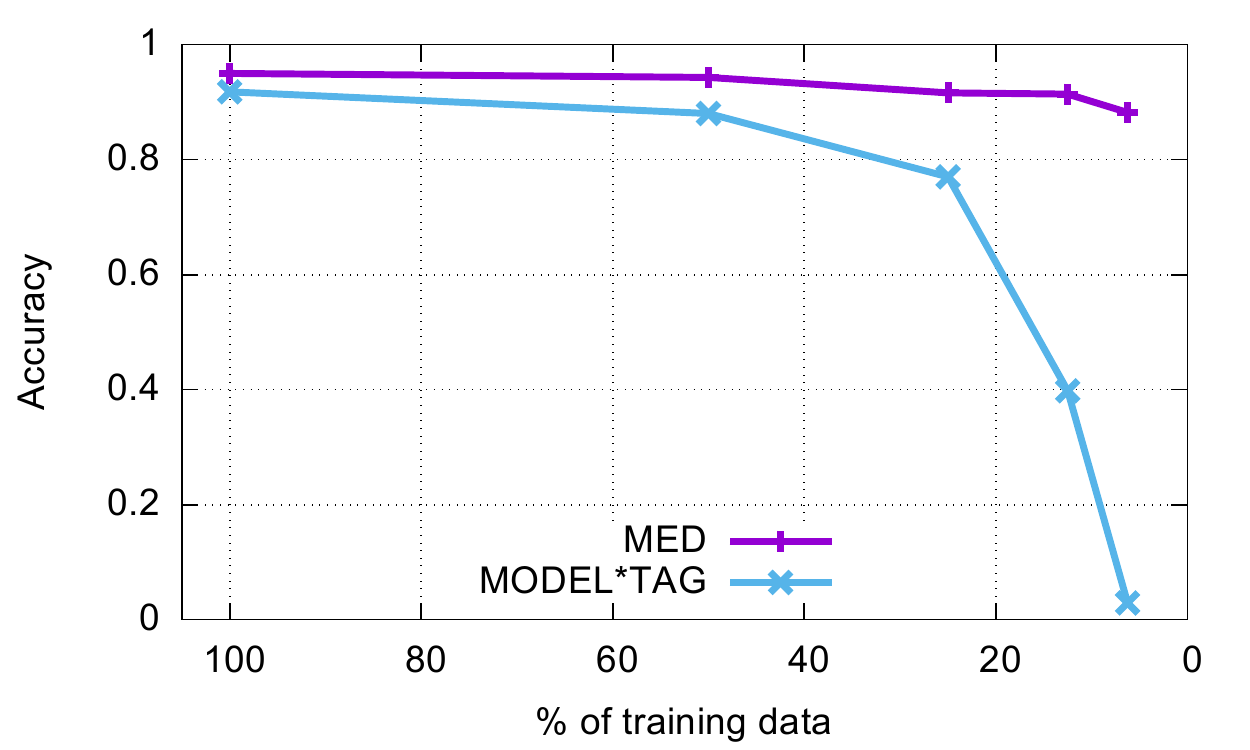}
  \caption{Results for the large general training set
    experiment: effect of reducing the training set \emph{for only
    2PIE $\rightarrow$ 13PKE} on the accuracy for 2PIE $\rightarrow$ 13PKE for MED and MODEL*TAG.}
  \label{fig:reductionOfOneTag}
\end{figure}

\newparagraph{Large general training set.} 
We iteratively halve the training data for 2PIE
$\rightarrow$ 13PKE until  only $6.25\%$ or $32$
samples are left.  
Figure \ref{fig:reductionOfOneTag} shows that MED
performs well even if only $6.25\%$ of the
training examples for the tag pair remain. In contrast, MODEL*TAG
struggles to generalize correctly. This is due to
the fact that we train one single model for all tags, so
it can learn from other tags and transfer what it has
learned to the tag pair that has a small training set.

\begin{figure}
  \centering
  \begin{minipage}[t]{0.23\textwidth}
  \includegraphics[width=\textwidth]{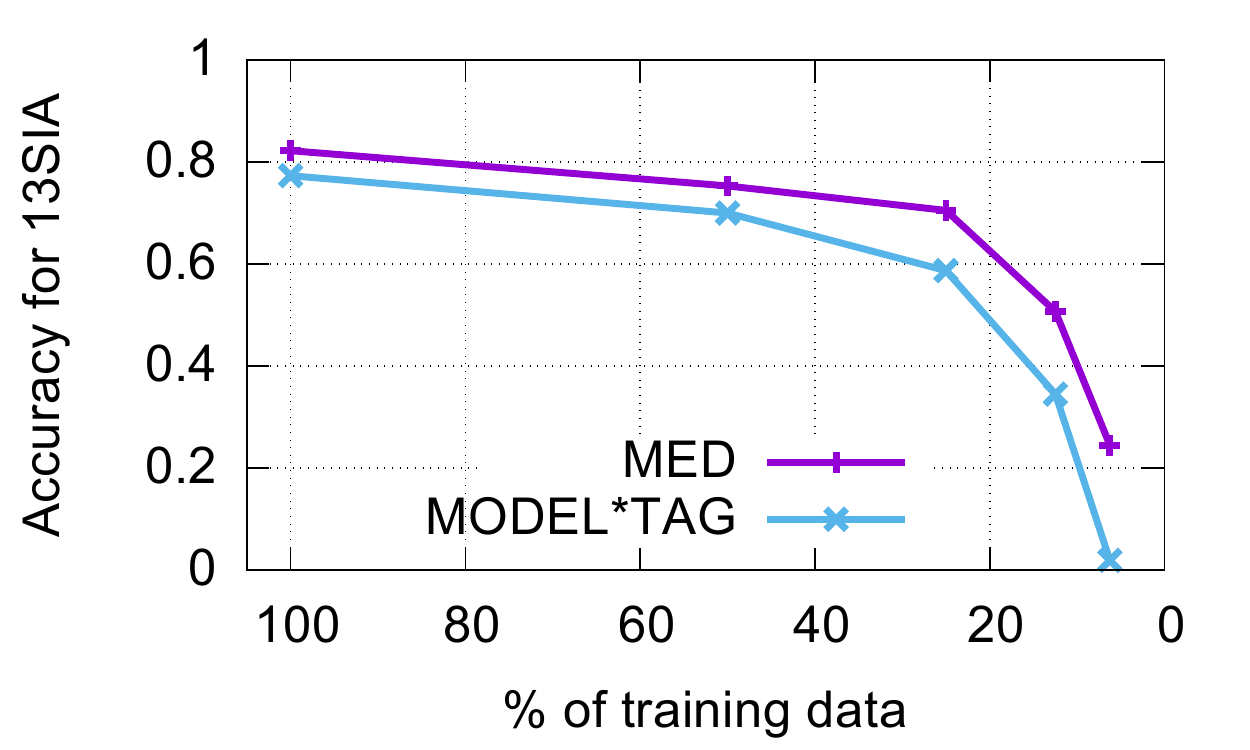}
  \end{minipage}
  \begin{minipage}[t]{0.23\textwidth}
  \includegraphics[width=\textwidth]{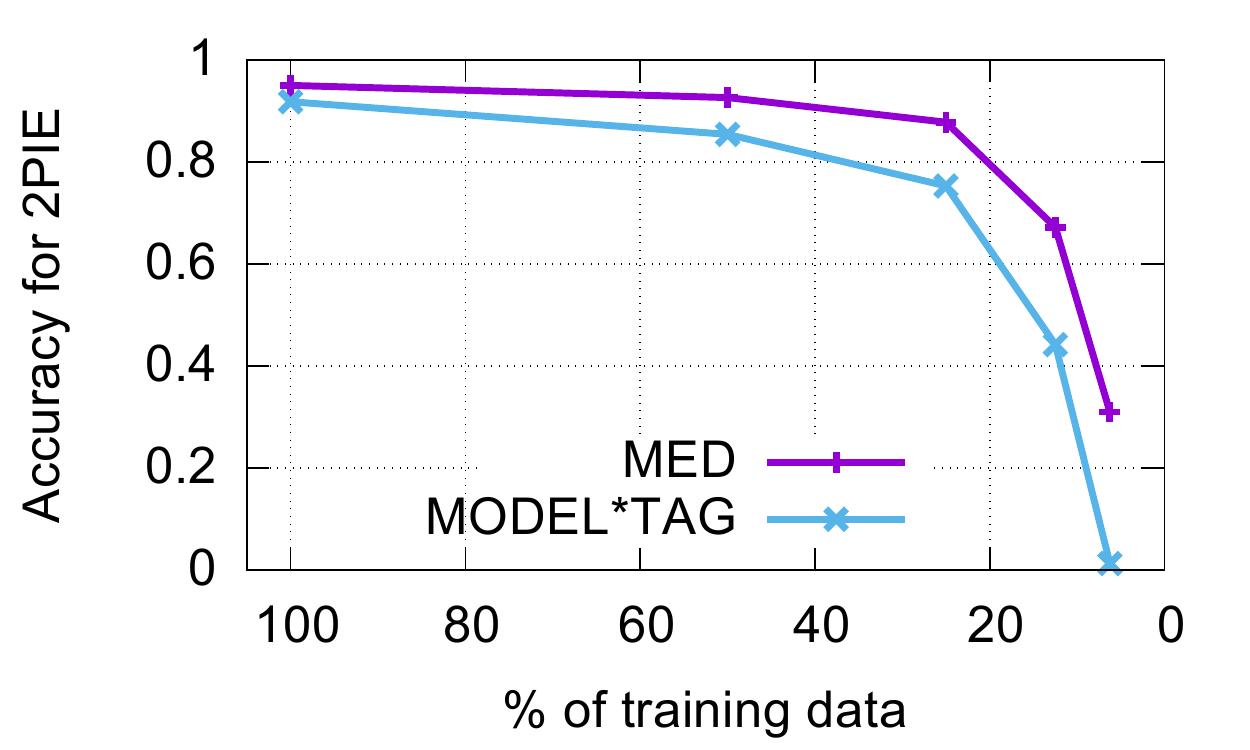}
  \end{minipage}
  \begin{minipage}[t]{0.23\textwidth}
  \includegraphics[width=\textwidth]{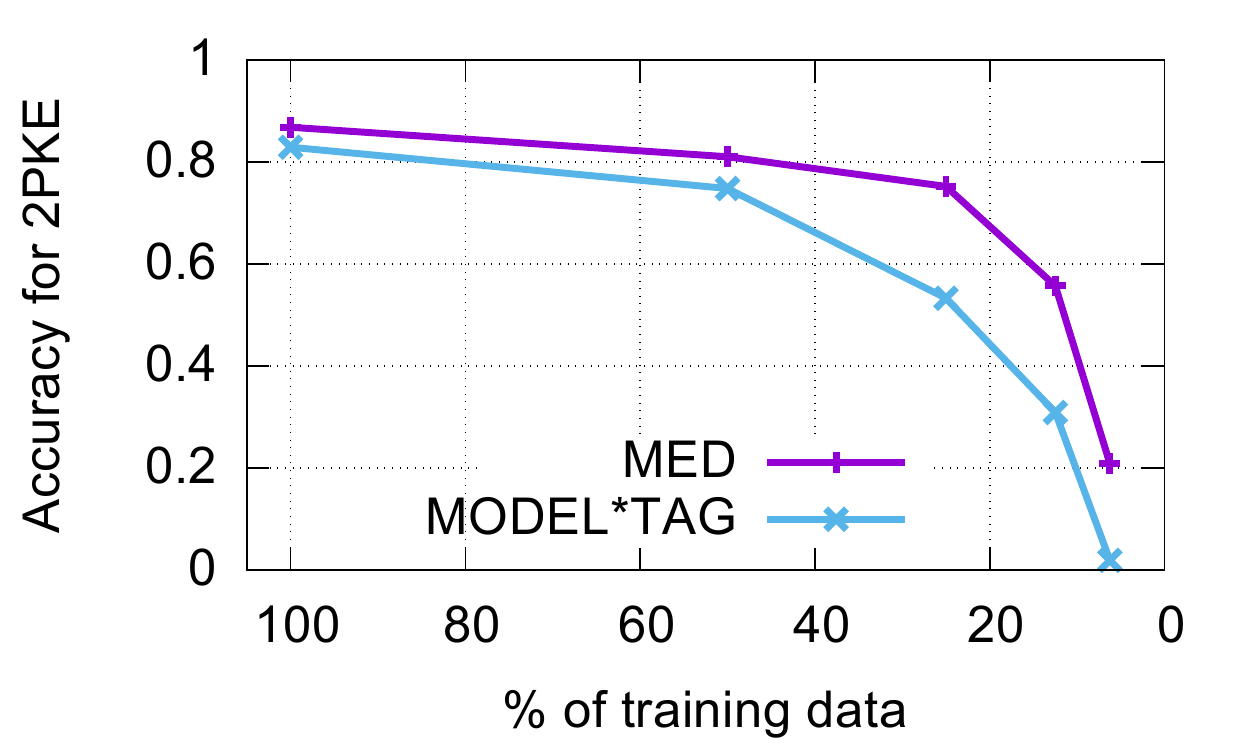}
  \end{minipage}
  \begin{minipage}[t]{0.23\textwidth}
  \includegraphics[width=\textwidth]{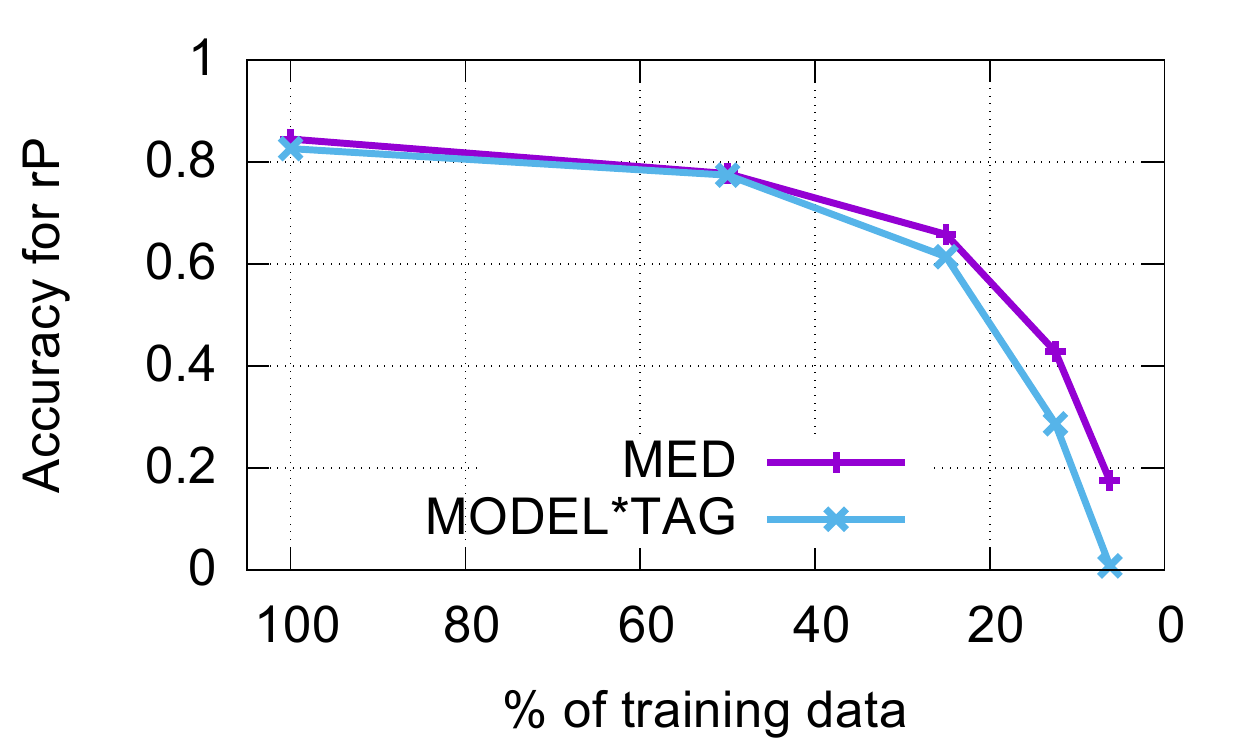}
  \end{minipage}
  \caption{Results for the small training set
    experiment: effect of reducing the training set \emph{for all
    tag pairs} on accuracy for MED and MODEL*TAG.}
  \label{fig:reductionAll}
\end{figure}

\newparagraph{Small training set.} 
Figure
\ref{fig:reductionAll} shows results for
reducing the training data
equally for all tags. MED performs much better
than the baseline for less than 50\% of the training
data. This can be explained by the fact that MED learns from all given data
at once and thus is able to learn common patterns that apply
across different
tag pairs.

\section{Related Work}
\label{sec:related_work}
Earlier work on morphology includes morphological segmentation \cite{harris1955,hafer1974word,dejean1998morphemes} and different approaches for MRI
\cite{ahlberg2014semi,durrett2013supervised,eskander2013automatic,nicolai2015inflection}.
\newcite{chrupala2008towards} defined edit trees and 
\newcite{chrupala2008towards} 
and \newcite{thomasjoint}
use
them for 
morphological tagging and lemmatization. 

In the last years, RNN encoder-decoder models and RNNs in general were applied to several NLP tasks.
For example, they proved to be useful for machine translation \cite{cho2014properties,sutskever2014sequence,bahdanau2014neural}, parsing \cite{vinyals2015grammar} and
speech recognition \cite{graves2005framewise,graves2013speech}.

MED bears some resemblance to
\newcite{DBLP:journals/corr/FaruquiTND15}'s work. However,
they train one network for every tag pair; this can
negatively impact performance for low-resource languages and
in general when training data are limited.  In contrast, we
train a single model for each language. This radically
reduces the amount of training data needed for the
encoder-decoder because most MRI patterns occur in many tag
pairs, so what is learned for one can be transferred to
others. To be able to model all tag pairs of the language together, we
introduce an explicit morphological representation that
enables the attention mechanism of the encoder-decoder to
generalize MRI patterns across tag pairs.

\section{Conclusion and Future Work}
\label{sec:conclusion}
We have presented MED, a language independent neural
sequence-to-sequence mapping approach, and POET, a method
based on edit trees for correcting the output of an MRI
system.  MED obtains results comparable to state-of-the-art
systems for CELEX and establishes the state-of-the-art for
SIGMORPHON16. POET improves results further for large
training sets. Our analysis showed that MED outperforms a
neural encoder-decoder baseline system by a large margin,
especially for small training sets.

In future work, we would like to make POET less dependent on
the source tag and thus increase its accuracy for small
training sets. Second, we will look into ways of taking
advantage of additional information sources including
unlabeled corpora.

\section*{Acknowledgments}
We gratefully acknowledge the financial support of Siemens
for this research.

\bibliography{acl2016}
\bibliographystyle{acl2016}

\end{document}